\documentclass{article}

\usepackage[nonatbib]{nips_2017}

\usepackage{nips_2017}

\usepackage[utf8]{inputenc} 
\usepackage[T1]{fontenc}    
\usepackage{hyperref}       
\usepackage{url}            
\usepackage{booktabs}       
\usepackage{amsfonts}       
\usepackage{nicefrac}       
\usepackage{microtype}      
\usepackage{graphicx}           
\usepackage{subfigure}
\usepackage{tikz}
\usepackage{url}
\usepackage{amsmath} 
\usepackage{bm}      
\usepackage[sectionbib]{chapterbib}
\usepackage[numbers]{natbib}

\title{Representation Learning on Large and Small Data}

\author{
  Chun-Nan Chou \\
  HTC AI Research, Taiwan \\
  jason.cn\_chou@htc.com \\
  \And
  Chuen-Kai Shie\\
  HTC AI Research, Taiwan \\
  chuenkai\_shie@htc.com \\
  \And
  Fu-Chieh Chang \\
  HTC AI Research, Taiwan \\
  mark.fc\_chang@htc.com \\
  \And 
  Jocelyn Chang \\
  Johns Hopkins University, USA \\
  jocelynjchang@gmail.com \\
  \And 
  Edward Y. Chang \\
  HTC AI Research, USA \\
  edward\_chang@htc.com \\
}

\begin{document}

\maketitle


\section{Introduction}
\label{dmd:introduction}

Extracting useful features from a scene is an essential step in any computer vision and multimedia data analysis task. Though progress has been made in past decades, it is still quite difficult for computers to comprehensively and accurately recognize an object or pinpoint the more complicated semantics of an image or a video. Thus, feature extraction is expected to remain an active research area in advancing computer vision and multimedia data analysis for the foreseeable future. \index{representation learning}

The approaches in feature extraction can be divided into two categories: {\em model-centric} and {\em data-driven}. 
The {model-centric} \index{model-centric} approach relies on human heuristics to develop a computer model (or algorithm) to extract features from an image. (We use imagery data as our example throughout this chapter.) Some widely used models are Gabor filter, wavelets, and SIFT~\cite{SIFT_IJCV04}. These models were engineered by scientists and then validated via empirical studies. A major shortcoming of the model-centric approach is that unusual circumstances that a model does not take into consideration during its design, such as different lighting conditions and unexpected environmental factors, can render the engineered features less effective.  Contrast to the model-centric approach, which dictates representations independent of data, the {data-driven} \index{data-driven} approach learns representations from data \cite{chang2011foundationsch2}. Example data-driven algorithms are multilayer perceptron (MLP) and convolutional neural network (CNN), which belong to the general category of neural network and deep learning \cite{hinton2007learning,hinton2006fast}.

Both model-centric and data-driven approaches employ a model (algorithm or machine). The differences between model-centric and data-driven can be told in two related aspects:
\begin{itemize}
\item {Can data affect model parameters?}
With model-centric, training data does not affect the model.  With data-driven, such as MLP or CNN, their internal parameters are changed/learned based on the discovered structure in large data sets \cite{DL-Nature2015}.
\item {Can data affect representations?} 
Whereas more data can help a data-driven approach to improve representations, more data cannot change the features extracted by a model-centric approach. For example, the features of an image can be affected by the other images in CNN (because the structure parameters modified through backpropagation are affected by all training images). But the feature set of an image is invariant of the other images in a model-centric pipeline such as SIFT.
\end{itemize}

The greater the quantity and diversity of data, the better the representations can be learned by a data-driven pipeline. In other words, if a learning algorithm has seen enough training instances of an object under various conditions, e.g., in different postures and has been partially occluded, then the features learned from the training data will be more comprehensive.


The focus of this chapter is on how neural network, specifically convolutional neural network (CNN), achieves effective representation learning. 
\textit{Neural network}, a neuro\-science-mo\-ti\-vat\-ed model, was based on Hubel and Wiesel's research on cats' visual cortex~\cite{hubel1968receptive}, and subsequently formulated into computation models by scientists in the early 1980s. Pioneer neural network models include Neocognitron \cite{fukushima1982neocognitron} and the shift-invariant neural network \cite{mcdermott1989shift}.  Widely cited enhanced models include LeNet-$5$ \cite{lecun1998gradient} and Boltzmann machines \cite{Hinton1986BZM}. 
However, popularity of neural networks
surged only in 2012 after large training 
data sets became available.
In 2012, \citet{krizhevsky2012imagenet} applied deep convolutional networks on the ImageNet dataset\footnote{ImageNet is a dataset of over $15$ million labeled images belonging to $22,000$ categories roughly \cite{deng2009imagenet}.} and their AlexNet achieved breakthrough accuracy in the ImageNet Large-Scale Visual Recognition Challenge (ILSVRC) 2012 competition\footnote{The ImageNet Large Scale Visual Recognition Challenge (ILSVRC) \cite{ILSVRC-site} evaluates algorithms for object detection and image classification on a subset of ImageNet, $1.2$ million images over $1,000$ categories. Throughout this chapter, we focus on discussing image classification challenge.}. This work convinced the research community and related industries that representation learning with big data is promising. Subsequently, several efforts aim to further improve learning capability of neural network. \index{neural networks} Today, the top-$5$ error rate\footnote{The top-$5$ error used to evaluate the performance of image classification is the proportion of images such that the ground-truth category is outside the top-$5$ predicted categories.} for the ILSVRC competition has dropped to $3.57\%$, a remarkable achievement considering the error rate was $26.2\%$ before AlexNet \cite{krizhevsky2012imagenet} was proposed.


We divide the remainder of this chapter into two parts, before suggesting related readings in concluding remarks.

The first part reviews representative CNN models proposed since 2012. These key representatives are discussed in terms of three aspects addressed in He's tutorial presentation \cite{he2016tutorial} at ICML 2016: $1$) representation ability, $2$) optimization ability, and $3$) generalization ability. The representation ability is the ability of a CNN to learn/capture representations from training data assuming the optimum could be found. Here, the optimum is referred to attaining the best solution of the underneath learning algorithm, modeled as an optimization problem. Specifically on CNN, the optimization problem is to find the optimal solution of stochastic gradient decent. This leads to the second aspect that He's tutorial addresses: the optimization ability.  The optimization ability is the feasibility of finding an optimum. Finally, the generalization ability is the quality of the test performance once model parameters have been learned from training data. 

The second part of this chapter deals with the small data problem. We present how features learned from one source domain with big data can be transferred to a different target domain with small data. This transfer representation learning approach is critical for remedying the small data challenge often encountered in the medical domain. \index{transfer learning} We use the Otitis Media detector, designed and developed for our XPRIZE Tricorder \cite{tricorder_website} device (code name DeepQ), to demonstrate how learning on a small dataset can be bolstered by transferring over learned representations from ImageNet, a dataset that is entirely irrelevant to otitis media.


\section{Representative Deep CNNs}
\label{sec-cnns}

Deep learning  roots in neuroscience. Strongly driven by the fact that the human visual system can effortlessly recognizing objects, neuroscientists have been developing vision models based on physiological evidences that can be applied to computers. Though such research may still be in its infancy and several hypotheses remain to be validated, some widely accepted theories have been established. Built upon the pioneer neuroscience work of Hubel~\cite{hubel1968receptive}, all recent models are founded on the theory that visual information is transmitted from the primary visual cortex (V1) over extrastriate visual areas (V2 and V4) to the inferotemporal cortex (IT).  IT in turn is a major source of input to the prefrontal cortex (PFC), which is involved in linking perception to memory and action~\cite{miller2000prefrontal}.

The pathway from V1 to IT, called the {\em ventral visual pathway}~\cite{kravitz2013ventral}, \index{ventral visual pathway} consists of a number of simple and complex layers. The lower layers detect simple features (e.g. oriented lines) at the pixel level. The higher layers aggregate the responses of these simple features to detect complex features at the object-part level. Pattern reading at the lower layers is unsupervised, whereas recognition at the higher layers involves supervised learning. 
Pioneer computational models developed based on the scientific evidences include Neocognitron \cite{fukushima1982neocognitron} and the shift-invariant neural network \cite{mcdermott1989shift}.  Widely cited enhanced models include LeNet-$5$ \cite{lecun1998gradient} and Boltzmann machines \cite{Hinton1986BZM}.  The remainder of this chapter uses representative CNN models, which stem from LeNet-$5$~\cite{lecun1998gradient}, to present three  design aspects: representation, optimization, and generalization.

CNNs are composed of two major components: {\em feature extraction} and {\em classification}. For feature extraction, a standard structure consists of stacked convolutional layers, which are followed by optional layers of contrast normalization or pooling. For classification, there are two widely used structures. One structure employs one or more fully-connected layers. The other structure uses a global average pooling layer, which is illustrated in Subsection~\ref{cnn:nin}. 

The accuracy of several computer vision tasks, such as that of house number recognition \cite{sermanet2012convolutional}, traffic sign recognition \cite{sermanet2011traffic}, and face recognition \cite{li2015convolutional}, has been substantially improved recently, thanks to advances in CNNs. For many similar object-recognition tasks, the superiority of CNNs over other methods is that CNNs join classification with feature extraction. 
Several works such as \cite{zeiler2014visualizing} show that CNNs can learn superior representations to boost the performance of  classification. 
Table~\ref{oc_table} shows four top-performing CNN models proposed over the past four years and their performance statistics in top-$5$ error rate. These representative models mainly differ in their number of layers or parameters. (Parameters refer to the learnable variables by supervised training including weight and bias parameters of the CNN models.)
Besides the four CNN models depicted in Table~\ref{oc_table}, \citet{lin2013network} proposed Network in Network (NIN), which has considerably influenced subsequent models such as GoogLeNet, vGG, and ResNet. In the following subsections, we thus present these five models' novel ideas and key techniques, which have had significant impacts on designing subsequent CNN models.

\begin{table}[t]
  \caption{Image classification performance on the ImageNet subset designated for ILSVRC \cite{ILSVRC-site}}
  \newcounter{counter1}
 \newcommand\rownumber{\stepcounter{counter1}\arabic{counter1}}
  \label{oc_table}
  \renewcommand\arraystretch{1.2}
  \centering
  \begin{tabular}{|l|lllll|}
    \toprule
    Model      & Year & Rank & Error    & \# of parameter &\# of parameters in\\
               &      &       & (top-$5$)  & layers           &a single model\\ 
    \midrule
    AlexNet \cite{krizhevsky2012imagenet}    & 2012 & $1$st   & $16.42\%$  & 8   &  $60$m \\ 
    VGG \cite{simonyan2014very}        & 2014 & $2$nd   &  $7.32\%$  & $19$  &  $144$m  \\
    GoogLeNet \cite{szegedy2015going}  & 2014 & $1$st   &  $6.67\%$  & $22$  & $5$m \\
    ResNet \cite{he2015deep}    & 2015 & $1$st   &  $3.57\%$  & $152$ & $60$m \\
 \hline   
 \bottomrule
  \end{tabular}
\end{table}

\subsection{AlexNet}
\label{cnn:alexnet}
\citet{krizhevsky2012imagenet} proposed AlexNet, which was the winner of the ILSVRC-$2012$ competition and outperformed the runner-up significantly (top-$5$ error rate of $16\%$ in comparison with $26\%$). The outstanding performance of AlexNet led to increased prevalence of CNNs in the computer vision field. AlexNet achieved this breakthrough performance by combining several novel ideas and effective techniques. Based on He's three aspects of learning deep models \cite{he2016tutorial}, these novel ideas and effective techniques can be categorized as follows:

\begin{enumerate}
\item {\em Representation ability}. In contrast to prior CNN models such as LetNet-$5$ ~\cite{lecun1998gradient}, AlexNet was deeper and wider in the sense that both the number of parameter layers and the number of parameters are larger than those of its predecessors.
\item {\em Optimization ability}. AlexNet utilized a non-saturating activation function, Rectified Linear Unit (ReLU) function, to make training faster.
\item {\em Generalization ability}. AlexNet employed two effective techniques, data augmentation and dropout, to alleviate overfitting.
\end{enumerate}

AlexNet's three key ingredients according to the description in \cite{krizhevsky2012imagenet} are Relu nonlinearity, data augmentation, and dropout.

\subsubsection{ReLU nonlinearity}
In order to model nonlinearity, the neural network introduces the activation function during the evaluation of neuron outputs. The traditional way to evaluate a neuron output $f$ as a function $g$ of its input $z$ is with $f = g(z)$ where $g$ can be a sigmoid function $g(z)=(1+e^{-z})^{-1}$ or a hyperbolic tangent function $g(z)=tanh(z)$. Both of these functions are saturating nonlinear. That is, the ranges of these two functions are each fixed between a minimum and maximum value.

\begin{figure}[h]
  \centering
  \includegraphics[scale=.45]{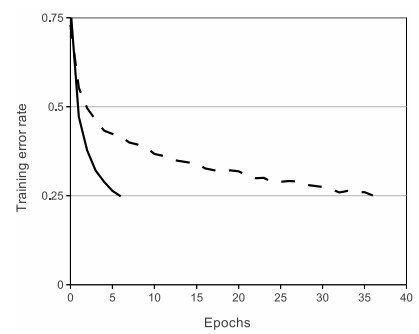}
  \caption{Six times faster in convergence with ReLU activation function compared to a hyperbolic tangent function from ~\cite{krizhevsky2012imagenet}}
  \label{relu_sixtimes}
\end{figure}

Instead of using saturating activation functions, however, AlexNet adopted the non-saturating activation function ReLU proposed in \cite{nair2010rectified}. ReLU computes the function $g(z)=max(0,z)$, which has a threshold at zero. Using ReLU enjoys two benefits. First, ReLU requires less computation in comparison with sigmoid and hyperbolic tangent functions, which involve expensive exponential operations. The other benefit is that ReLU, in comparison to sigmoid and hyperbolic tangent functions, is found to accelerate the convergence of stochastic gradient descent (SGD). Reproduced from \cite{krizhevsky2012imagenet}, Fig.~\ref{relu_sixtimes} shows that the CNN with ReLU is six times faster to train than that with a hyperbolic tangent function. Due to these two advantages, recent CNN models have adopted ReLU as their activation functions.

\subsubsection{Data augmentation}
As shown in Table~\ref{oc_table}, the AlexNet architecture has $60$ million parameters. This huge number of parameters makes overfitting highly possible if raining data is not sufficient. To combat overfitting, AlexNet incorporated two primary constructs: data augmentation and dropout.

Thanks to ImageNet, AlexNet is the first model that enjoys {\em big data} and takes advantage of benefits from the data-driven feature learning approach advocated by \cite{chang2011foundationsch2}.  However, even the $1.2$ million ImageNet labeled instances are still considered insufficient given that the number of parameters is $60$ million. (From simple algebra, $1.2$ million equations are insufficient for solving $60$ million variables.) Conventionally, when the training dataset is limited, the common practice in image data is to artificially enlarge the dataset by using label-preserving transformations \cite{ciregan2012multi,cirecsan2011high,simard2003best}. In order to enlarge the training data, AlexNet employed two distinct forms of data augmentation, both of which can produce the transformed images from the original images with very little computation \cite{krizhevsky2012imagenet,li2003discovery}.

The first scheme of data augmentation includes a random cropping function and horizontal reflection function. Data augmentation can be applied to both the training and testing stages. For the training stage, AlexNet randomly extracted smaller image patches $(224 \times 224)$ and their horizontal reflections from the original images $(256 \times 256)$. The AlexNet model was trained on these extracted patches instead of the original images in the ImageNet dataset. In theory, this scheme is capable of increasing the training data by a factor of $(256-224) \times (256-224) \times 2 = 2,048$. Although the resultant training examples are highly interdependent, \cite{krizhevsky2012imagenet} claimed that without this data augmentation scheme, the AlexNet model would suffer from substantial overfitting. (This is evident from our algebra example.) For the testing stage, AlexNet generated ten patches, including four corner patches, one center patch, and each of the five patches' horizontal reflections from test images. Based on the generated ten patches, AlexNet firstly derived temporary results from the network’s softmax layer and secondly made a prediction by averaging the ten results.

The second scheme of data augmentation alters the intensities of the RGB channels in training images by using principal component analysis (PCA). This scheme is used to capture an important property of natural images: the invariance of object identity to changes in the intensity and color of the illumination. The detailed implementation is as follows. First, the principal components of RGB pixel values are acquired by performing PCA on a set of RGB pixel values throughout the ImageNet training set. When a particular training image is chosen to train the network, each RGB pixel $I_{xy} = [I_{xy}^{R}, I_{xy}^{G}, I_{xy}^{B}]^{T}$ of this chosen training image is refined by adding the following quantity:
\[
[\mathbf{\beta}_{1}, \mathbf{\beta}_{2}, \mathbf{\beta}_{3}][\alpha_{1}\lambda_{1}, \alpha_{2}\lambda_{2}, \alpha_{3}\lambda_{3}]^{T},
\]
where $\mathbf{\beta}_{i}$ and $\lambda_{i}$ represent the $i^{th}$ eigenvector and eigenvalue of the $3 \times 3$ covariance matrix of RGB pixel values respectively, and $\alpha_{i}$ is a random variable drawn from a Gaussian model with mean zero and standard deviation $0.1$. Please note that, each time one training image is chosen to train the network, each $\alpha_{i}$ is re-drawn. Thus, during the training, $\alpha_{i}$ of data augmentation varies with different times for the same training image. Once $\alpha_{i}$ is drawn, $\alpha_{i}$ is applied to all the pixels of this chosen training image.

\subsubsection{Dropout}
Model ensembles such as {\em bagging} \cite{breiman1996bagging}, {\em boosting} \cite{freund1995desicion}, and {\em random forest} \cite{breiman2001random} have long been shown to  effectively reduce class-prediction variance and hence testing error. Model ensembles rely on combing the predictions from several different models. However, this method is impractical for large-scale CNNs such as AlexNet, since training even one CNN can take several days or even weeks.

Rather than training multiple large CNNs, \citet{krizhevsky2012imagenet} employed the “dropout” technique introduced in \cite{hinton2012improving} to efficiently perform model combination. This technique simply sets the output of each hidden neuron to zero with a probability $\mathbf{p}$ (e.g. $0.5$ in \cite{krizhevsky2012imagenet}). Afterwards, the dropped-out neurons neither contribute to the forward pass nor participates in the subsequent back-propagation pass. In this manner, different network architectures are sampled when each training instance is presented, but all these sampled architectures share the same parameters. In addition to combining models efficiently, the drop-out technique has the effect of reducing the complex co-adaptations of neurons, since a neuron cannot depend on the presence of other neurons. In this way, more robust features are forcibly learned. At the time of testing, all neurons are used, but their outputs are multiplied by $\mathbf{p}$, which is a reasonable approximation of the geometric mean of the predictive distributions produced by the exponential quantity of dropout networks \cite{hinton2012improving}.

In \cite{krizhevsky2012imagenet}, dropout was only applied to the first two fully-connected layers of AlexNet and roughly doubled the number of iterations required for convergence. \citet{krizhevsky2012imagenet} also claimed that AlexNet suffered from substantial overfitting without dropout.

\subsection{NIN}
\label{cnn:nin}
Although NIN, presented in \cite{lin2013network}, has not ranked among the best of ILSVRC competitions in recent years, its novel designs have significantly influenced subsequent CNN models, especially its $1 \times 1$ convolutional filters. The $1 \times 1$ convolutional filters are widely-used by current CNN models and have been incorporated into VGG, GoogLeNet, and ResNet. Based on He's three aspects of learning deep models, the novel designs proposed in NIN can be categorized as follows:

\begin{enumerate}
\item {\em Representation ability}. In order to enhance the model's discriminability, NIN adopted multilayer perceptron (MLP) convolutional layers with more complex structures to abstract the data within the receptive field.
\item {\em Optimization ability}.  Optimization in NIN remained typical compared to that of the other models.
\item {\em Generalization ability}. NIN utilized global average pooling over feature maps in the classification layer, because global average pooling is less prone to overfitting than traditional fully-connected layers.
\end{enumerate}

\subsubsection{MLP convolutional layer}

The work of \citet{lin2013network} argued that the conventional CNNs \cite{lecun1998gradient} implicitly make the assumption that the samples of the latent concepts are linearly separable. Thus, typical convolutional layers generate feature maps with linear convolutional filters followed by nonlinear activation functions. This kind of feature maps can be calculated as follows:

\begin{equation}
f_{i,j,k} = g(z_{i,j,k}),\ {\it where}\ z_{i,j,k} = w^{T}_{k}x_{i,j}+b_{k}.
\end{equation}
Here $(i,j)$ is the pixel index, and $k$ is the filter index. Parameter $x_{i,j}$ stands for the input patch centered at location $(i,j)$. Parameters $w_{k}$ and $b_{k}$ represent the weight and bias parameters of the $k$-th filter respectively. Parameter $z$ denotes the result of the convolutional layer and the input to the activation function, while $g$ denotes the activation function, which can be a sigmoid $g(z)=(1+e^{-z})^{-1}$, hyperbolic tangent $g(z)=tanh(z)$, or ReLU $g(z)=max(z,0)$.

\begin{figure}
\centering
\subfigure[Linear convolutional layer]{\centering\includegraphics[width=.3\linewidth]{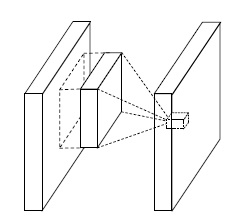}}
\hspace{0.1\textwidth}%
\subfigure[Mlpconv layer]{\centering\includegraphics[width=.39\linewidth]{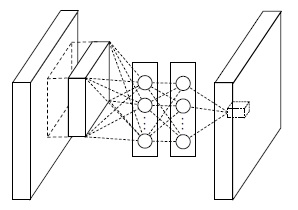}}
\caption{Comparison of linear convolutional layer and mlpconv layer from \cite{lin2013network}}
\label{fig:nin_mlpconv}
\end{figure}

However, instances of the same concept often live on a nonlinear manifold. Hence, the representations that capture these concepts are generally highly nonlinear functions of the input. In NIN, the linear convolutional filter is replaced with an MLP. This new type of layer is called mlpconv in \cite{lin2013network}, where MLP convolves over the input. There are two reasons for choosing an MLP. First, an MLP is a general nonlinear function approximator. Second, an MLP can be trained by using back-propagation, and is therefore compatible with conventional CNN models. Fig.~\ref{fig:nin_mlpconv} depicts the difference between a linear convolutional layer and an mlpconv layer. The calculation for an mlpconv layer is performed as follows:

\begin{equation}
f_{i,j,k_{1}}^1=g(z_{i,j,k_{1}}^1),\ {\it where}\ z_{i,j,k_{1}}^1 = {w_{k_{1}}^{1}}^Tx_{i,j}+b_{k_{1}}^1
\end{equation}
\begin{equation}
f_{i,j,k_{2}}^2=g(z_{i,j,k_{2}}^2),\ {\it where}\ z_{i,j,k_{2}}^2 = {w_{k_{2}}^{2}}^Tf_{i,j}^{1}+b_{k_{2}}^2
\label{eq:11conv}
\end{equation}
\[
\vdots
\]
\begin{equation}
f_{i,j,k_{n}}^n=g(z_{i,j,k_{n}}^n),\ {\it where}\ z_{i,j,k_{n}}^n = {w_{k_{n}}^{n}}^Tf_{i,j}^{n-1}+b_{k_{n}}^n
\end{equation}
Here, $n$ is the number of layers in the MLP, and $k_{i}$ is the filter index of the $i^{th}$ layer. \citet{lin2013network} used ReLU as the activation function in the MLP.

From a pooling point of view, Eq.~\ref{eq:11conv} is equivalent to performing cross channel parametric pooling on a typical convolutional layer. Traditionally, there is no learnable parameter involved in the pooling operation. Besides, the conventional pooling is performed within one particular feature map, and is thus not a cross channel. However, Eq. \ref{eq:11conv} performs a weighted linear recombination on the input feature maps, which then goes through a nonlinear activation function. Therefore, \citet{lin2013network} interpreted Eq. \ref{eq:11conv} as a cross channel parametric pooling operation. They also suggested that we can view Eq. \ref{eq:11conv} as a convolutional layer with a $1 \times 1$ convolutional filter.

\subsubsection{Global average pooling}

\citet{lin2013network} made the following remarks. The traditional CNN adopts the fully-connected layers for classification. Specifically, the feature maps of the last convolutional layer are flattened into a vector, and this vector is fed into some fully-connected layers followed by a softmax layer \cite{krizhevsky2012imagenet,zeiler2013stochastic,goodfellow2013maxout}. In this fashion, convolutional layers are treated as feature extractors, using traditional neural networks to classify the resulting features. However, the traditional neural networks are prone to overfitting, thereby degrading the generalization ability of the overall network.

\begin{figure}[h]
  \centering
  \includegraphics[scale=.6]{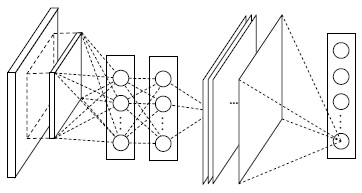}
  \caption{Global average pooling layer from ~\cite{lin2013network}}
  \label{fig:global_pooling}
\end{figure}

Instead of using the fully-connected layers with regularization methods such as dropout, \citet{lin2013network} proposed global average pooling to replace the traditional fully-connected layers in CNNs. Their idea was to derive one feature map from the last mlpconv layer for each corresponding category of the classification task. The values of each derived feature map would be averaged spatially, and all the average values would be flattened into a vector which would then be fed directly into the softmax layer. Fig.~\ref{fig:global_pooling} delineates the design of global average pooling. One advantage of global average pooling over fully-connected layers is that there is no parameter to optimize in global average pooling, preventing overfitting at this layer. Another advantage is that the linkage between feature maps of the last convolutional layer and categories of classification can be easily interpreted, which allows for better understanding. Finally, global average pooling aggregates spatial information and thus offers more robust spatial translations of the input.

\subsection{VGG}
\label{cnn:vgg}
VGG, proposed by \citet{simonyan2014very}, ranked first and second in the localization and classification tracks of the ImageNet Challenge 2014, respectively. VGG reduced the top-$5$ error rate of AlexNet from $16.42\%$ to $7.32\%$, which is an improvement of more than $50\%$. Using very small ($3 \times 3$) convolutional filters makes a substantial contribution to this improvement. Consequently, very small ($3 \times 3$) convolutional filters have been very popular in recent CNN models. Here, the convolutional filter is small or large, depending on the size of its receptive field. According to He's three aspects of learning deep models, the essential ideas in VGG can be depicted as follows:
\begin{enumerate}
\item {\em Representation ability}. VGG used very small ($3 \times 3$) convolutional filters, which make the decision function more discriminative. Additionally, the depth of VGG was increased steadily to $19$ parameter layers by adding more convolutional layers, an increase that is feasible due to the use of very small ($3 \times 3$) convolutional filters in all layers.
\item {\em Optimization ability}. VGG used very small ($3 \times 3$) convolutional filters, thereby decreasing the number of parameters.
\item {Generalization ability}. VGG employed multi-scale training to recognize objects over a wide range of scales.
\end{enumerate}

\subsubsection{Very small convolutional filters}

According to \cite{simonyan2014very}, instead of using relatively large convolutional filters in the first convolutional layers (e.g. $11 \times 11$ with stride 4 in \cite{krizhevsky2012imagenet} or $7 \times 7$ with stride 2 in \cite{zeiler2014visualizing,sermanet2013overfeat}), VGG used very small $3 \times 3$ convolutional filters with stride 1 throughout the whole network. The output dimension of a stack of two $3 \times 3$ convolutional filters (without spatial pooling operation in between) is equal to the output dimension of one $5 \times 5$ convolutional filter. Thus, \cite{simonyan2014very} claimed that a stack of two $3 \times 3$ convolutional filters has an effective receptive field of $5 \times 5$. By following the same rule, we can conclude that three such filters construct a $7 \times 7$ effective receptive field.

The reasons for using smaller convolutional filters are twofold. Firstly, the decision function is more discriminative. For example, using a stack of three $3 \times 3$ convolutional filters instead of a single $7 \times 7$ convolutional filter can incorporate three nonlinear activation functions instead of using just one. Secondly, the number of parameters can be decreased. Assuming that the input as well as output feature maps have $\mathbf{C}$ channels, we can use our prior example as an illustration of decreased parameter number. The stack of three $3 \times 3$ convolutional filters is parametrized by $3(3^2\mathbf{C}^2) = 27\mathbf{C}^2$ weight parameters. On the other hand, a single $7 \times 7$ convolutional filter requires $7^2\mathbf{C}^2 = 49\mathbf{C}^2$ weight parameters, which is 81\% more than that of three $3 \times 3$ filters. \citet{simonyan2014very} argued that we can view the usage of very small convolutional filers as imposing a regularization on the $7 \times 7$ convolutional filters and forcing them to have a decomposition through $3 \times 3$ filters (with nonlinearity injected in between).

\subsubsection{Multi-scale training}
\citet{simonyan2014very} considered two approaches for setting the training scale to $\mathbf{S}$. The first approach is to fix $\mathbf{S}$, which corresponds to single-scale training. The single-scale training has been widely used in prior art \cite{krizhevsky2012imagenet,zeiler2014visualizing,sermanet2013overfeat}. However, objects in images can be of different sizes, and it is beneficial to take objects of different sizes into account during the training phrase. Thus, the second approach proposed in VGG for setting to $\mathbf{S}$ is multi-scale training. In multi-scale training, each training image is individually rescaled by randomly sampling $\mathbf{S}$ from a certain range $[\mathbf{S}_{min}, \mathbf{S}_{max}]$. In VGG, $\mathbf{S}_{min}$ and $\mathbf{S}_{max}$ were set to 256 and 512 respectively. \citet{simonyan2014very} also interpreted this multi-scale training as a sort of data augmentation of the training set with scale jittering, where a single model is trained to recognize objects over a wide range of scales.

\subsection{GoogLeNet}
\label{cnn:googlenet}
GoogLeNet, devised by \citet{szegedy2015going}, held the record for classification and detection of ILSVRC 2014. GoogLeNet reached a top-5 error rate of $6.67\%$, which is better than that of VGG with $7.32\%$ in the same year. This improvement is mainly attributed to the proposed ``Inception module.'' According to He's three aspects of learning deep models, the essential ideas of GoogLeNet can be categorized as follows:

\begin{enumerate}
\item {\em Representation ability}. GoogLeNet increased the depth and width of the network while keeping the computational budget constant. Here, the depth and width of the network represent the number of network layers and the number of neurons at each layer, respectively.
\item {\em Optimization ability}. GoogLeNet improved utilization of computing resources inside the network through dimension reduction, thereby easing the training of networks.
\item {\em Generalization ability}. given the number of labeled examples in the training set is the same, GoogLeNet utilized dimension reduction to decrease the number of parameters dramatically and was hence less prone to overfitting.
\end{enumerate}

\subsubsection{Inception modules}
The main idea of GoogLeNet is to consider how an optimal local sparse structure of a CNN can be approximated and covered by readily available dense components. After this structure is acquired, all we need to do is to repeat it spatially. \citet{szegedy2015going} crafted the ``Inception module'' for the optimal local sparse structure. In the following, we explain the design principle of the Inception module according to \cite{szegedy2015going}.

Each neuron from a layer corresponds to some region of the input image, and these neurons are grouped into feature maps according to their common properties. In the lower layers (the layers closer to the input), the correlated neurons would concentrate on the same local region. Thus, we would end up with a lot of groups concentrated in a single region, and these groups can be covered by using $1 \times 1$ convolutional filters, as suggested in \cite{lin2013network}, justifying the use of $1 \times 1$ convolutional filters in the Inception module.

However, there may be a small number of groups that are more spatially spread out and thus require larger convolutional filters for coverage over the larger patches. Consequently, the size of the convolutional filters used depends on the size of its receptive field. In general, there will be a decreasing number of groups over larger and larger regions. In order to avoid patch-alignment issues, the larger convolutional filters of the Inception module are restricted to $3 \times 3$ and $5 \times 5$, a decision based more on convenience than on necessity.

\begin{figure}[h]
  \centering
      \tikzstyle{gnet_layer} = [draw=black, thick, minimum width=2.5cm, minimum height = 1.2cm, text width=2.5cm, align=center]
\begin{tikzpicture}[scale=.8]
\node[gnet_layer] (n1) {Feature Map \\ Concatenation};
\node[gnet_layer] (n2) [below of = n1,node distance=2cm]{3x3 \\ Convolutional Filters};
\node[gnet_layer] (n21) [left of = n2,node distance=3cm]{1x1 \\ Convolutional Filters};
\node[gnet_layer] (n22) [right of = n2,node distance=3cm]{5x5 \\ Convolutional Filters};
\node[gnet_layer] (n23) [right of = n22,node distance=3cm]{3x3 \\ Max Pooling};
\node[gnet_layer] (n3) [below of = n2,node distance=2cm]{ Previous \\ Feature Maps};
\draw[->,thick] (n3.north) -- (n2.south);
\draw[->,thick] (n3.north) -- (n21.south);
\draw[->,thick] (n3.north) -- (n22.south);
\draw[->,thick] (n3.north) -- (n23.south);
\draw[->,thick] (n2.north) -- (n1.south);
\draw[->,thick] (n21.north) -- (n1.south);
\draw[->,thick] (n22.north) -- (n1.south);
\draw[->,thick] (n23.north) -- (n1.south);
\end{tikzpicture}
  \caption{Naive version of the Inception module from \cite{szegedy2015going}}
  \label{fig:inception_naive}
\end{figure}
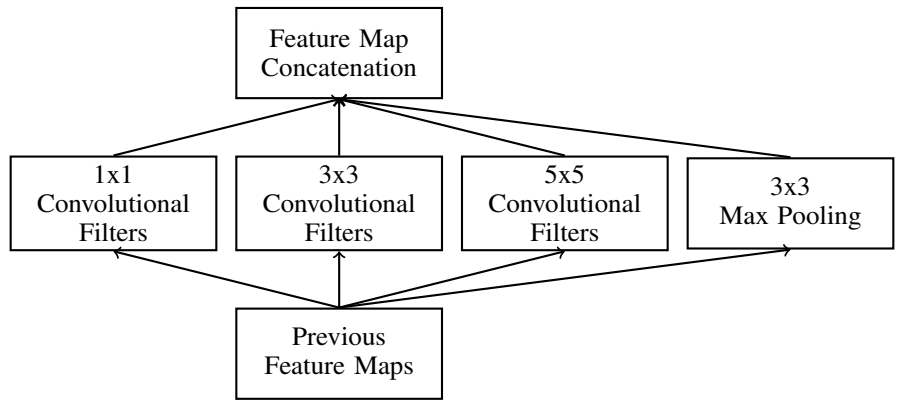

Additionally, since max pooling operations have been essential for the success of current CNNs, \cite{szegedy2015going} suggested that adding an alternative parallel pooling path in the Inception module could have additional beneficial effects. The Inception module is a combination of all aforementioned components including $1 \times 1$, $3 \times 3$, and $5 \times 5$ convolutional filters as well as $3 \times 3$ max pooling. Finally, their output feature maps are concatenated into a single output vector, forming the input for the next stage. Fig.~\ref{fig:inception_naive} shows the overall architecture of the devised Inception module.

\subsubsection{Dimension reduction}
However, as illustrated in \cite{szegedy2015going}, the devised Inception module introduces one big problem: even a modest number of $5 \times 5$ convolutional filters can be prohibitively expensive on top of a convolutional layer with a large number of feature maps. This problem becomes even more pronounced once max pooling operations get involved since the number of output feature maps equals the number of feature maps in the previous layer. The merging of outputs of the pooling operation with outputs of convolutional filters would lead to an inevitable increase in the number of feature maps from layer to layer. Although the devised Inception module might cover the optimal sparse structure, it would do so very inefficiently, leading possibly to a computational blow-up within a few layers \cite{szegedy2015going}.

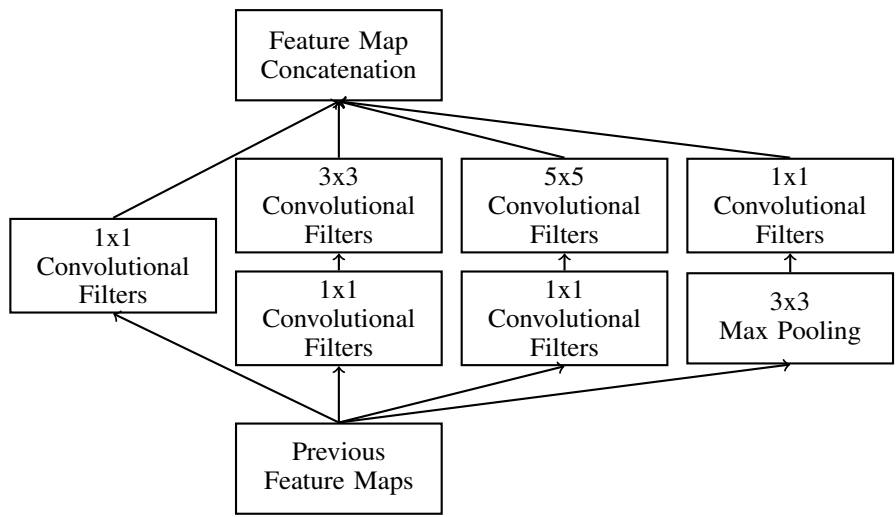
\begin{figure}[h]
  \centering
    \tikzstyle{gnet_layer2} = [draw=black, thick, minimum width=2.5cm, minimum height = 1.2cm, text width=2.5cm, align=center]
\begin{tikzpicture}[scale=.8]
\node[gnet_layer2] (n1) {Feature Map \\ Concatenation};
\node[gnet_layer2] (n2) [below of = n1,node distance=2cm]{3x3 \\ Convolutional Filters};
\node[gnet_layer2] (n22) [right of = n2,node distance=3cm]{5x5 \\ Convolutional Filters};
\node[gnet_layer2] (n23) [right of = n22,node distance=3cm]{1x1 \\ Convolutional Filters};
\node[gnet_layer2] (n3) [below of = n2,node distance=1.5cm]{ 1x1 \\ Convolutional Filters};
\node[gnet_layer2] (n32) [right of = n3,node distance=3cm]{ 1x1 \\ Convolutional Filters};
\node[gnet_layer2] (n33) [right of = n32,node distance=3cm]{ 3x3 \\ Max Pooling};
\coordinate [below of=n2,node distance=0.8cm] (c1) {};
\node[gnet_layer2] (n31) [left of = c1,node distance=3cm]{1x1 \\ Convolutional Filters};
\node[gnet_layer2] (n4) [below of = n3,node distance=2cm]{Previous \\ Feature Maps};
\draw[->,thick] (n3.north) -- (n2.south);
\draw[->,thick] (n32.north) -- (n22.south);
\draw[->,thick] (n33.north) -- (n23.south);
\draw[->,thick] (n2.north) -- (n1.south);
\draw[->,thick] (n22.north) -- (n1.south);
\draw[->,thick] (n23.north) -- (n1.south);
\draw[->,thick] (n4.north) -- (n31.south);
\draw[->,thick] (n4.north) -- (n32.south);
\draw[->,thick] (n4.north) -- (n33.south);
\draw[->,thick] (n4.north) -- (n3.south);
\draw[->,thick] (n31.north) -- (n1.south);
\end{tikzpicture}
  \caption{The Inception module with dimension reduction from \cite{szegedy2015going}.}
  \label{fig:inception_reduction}
\end{figure}

This dilemma inspired the second idea of the Inception module: to reduce dimensions judiciously only when the computational requirements would otherwise increase too much. For example, $1 \times 1$ convolutional filters are used to compute reductions before the more expensive $3 \times 3$ and $5 \times 5$ convolutional filters are used. In such a way, the number of neurons at each layer can be increased significantly without an uncontrolled blow-up in computational complexity at later layers. In addition to reductions, the Inception module also includes the use of ReLU activation functions for increased discriminative qualities. The final design is depicted in Fig.~\ref{fig:inception_reduction}.

\subsection{ResNet}
\label{cnn:resnet}
ResNet, proposed by \citet{he2015deep}, created a sensation in 2015 as the winner of several vision competitions in ILSVRC and COCO 2015, including ImageNet classification, ImageNet detection, ImageNet localization, COCO detection, and COCO segmentation. ResNet achieved a $3.57\%$ top-$5$ error rate on the ImageNet test set, which was an almost $50\%$ improvement from the 2014 winner, GoogLeNet, with a $6.67\%$ top-$5$ error rate. Residual learning plays a critical role in ResNet since it eases the training of networks, and the networks can gain accuracy from considerably increased depth. As reported in He's tutorial presentation \cite{he2016tutorial} at ICML 2016, ResNet addresses the three aspects of learning deep models as follows:

\begin{enumerate}
\item {\em Representation ability}. Although ResNet presents no explicit advantage on representation, it allowed models to go substantially deeper by re-parameterizing the learning between layers.
\item {\em Optimization ability}. ResNet enabled very smooth forward and backward propagation and hence greatly eased optimizing deeper models.
\item {\em Generalization ability}. Generalization is not explicitly addressed in ResNet, but \cite{he2016tutorial} argued that deeper and thinner models have good generalization.
\end{enumerate}

\subsubsection{Residual learning}
Motivated by the degradation of training accuracy in deeper networks \cite{he2015convolutional,srivastava2015highway}, the idea of residual learning was proposed in \cite{he2015deep} and was employed in ResNet. In accordance with \cite{he2015deep}, we will illustrate the idea of residual learning in the following.  Residual learning reformulates the few stacked layers as learning residual mappings with reference to the layer inputs, instead of learning unreferenced mappings.

Formally, let us denote $\boldsymbol{M}_{d}(f)$ as a desired underlying mapping to be fit to a few stacked layers, with $f$ denoting the inputs to the first of these layers. If one hypothesizes that multiple nonlinear layers can asymptotically approximate any complicated mapping, then it follows naturally to hypothesize that such layers can asymptotically approximate a residual mapping, i.e., $\boldsymbol{M}_{d}(f) - f$ (assuming that the input and output are of the same dimensions). Thus, rather than expecting stacked layers to approximate $\boldsymbol{M}_{d}(f)$, residual learning explicitly makes these layers approximate residual mapping $\boldsymbol{M}_{r}(f) := \boldsymbol{M}_{d}(f) - f$. The original mapping becomes $\boldsymbol{M}_{r}(f) + f$. Although learning both residual mappings and unreferenced mappings should enable asymptotic approximates of desired mappings (as hypothesized), the ease of learning might be different.

With residual learning reformulation, if identity mappings are optimal, the solvers may simply drive the parameters of the multiple nonlinear layers towards zero to approach identity mappings. Though identity mappings are unlikely to be optimal in real cases, the residual learning reformulation may help precondition the learning problem. If the optimal mapping is closer to an identity mapping than to a zero mapping, it should be easier for the solvers to find the perturbations with reference to an identity mapping than to learn the function as a new one.

\subsubsection{Identity mapping by shortcuts}

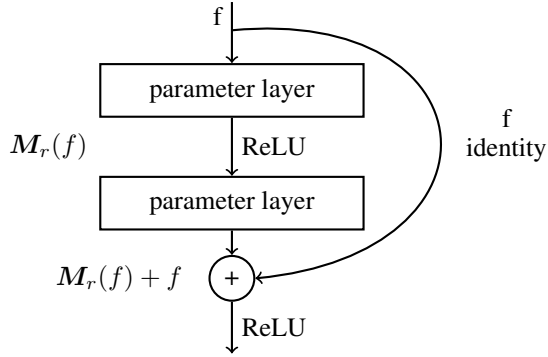
\begin{figure}[h]
  \centering
  \tikzstyle{res_layer} = [draw=black, thick, minimum width=3.5cm, minimum height = 0.7cm]
\begin{tikzpicture}
\node[minimum size=0] (n0) {}; 
\coordinate [below of=n0,node distance=0.5cm] (n1) {}; 
\node[res_layer] (l1) [below of=n1,node distance=0.8cm]{parameter layer};
\node[res_layer] (l2) [below of=l1,node distance=1.5cm]{parameter layer};
\coordinate [right of=l1,node distance=2.5cm](ll1) ;
\coordinate [right of=l2,node distance=2.5cm](ll2);
\node[circle, draw=black,thick](c1)[below of=l2,node distance=1cm]{+} ;
\coordinate [below of=c1,node distance=1.0cm] (n2) {};
\draw[->,thick] (n0) -- node[auto,left] {f} (n1) -- (l1);
\draw[->,thick] (l1.south) --node[auto,right](p1) {ReLU}  (l2.north);
\draw[->,thick] (l2.south) -- (c1.north);
\draw[->,thick] (c1.south) -- node[auto,right] {ReLU} (n2)  ;
\draw[->,thick] (n1.east) to [bend left=90,out=90,in=90, distance=3.5cm] node[auto,right] {\begin{tabular}{c} f \\ identity  \end{tabular}} (c1.east);
\node[auto,left of=p1,node distance=3.0cm] {$\boldsymbol{M}_{r}(f)$};
\node[auto,left of=c1,node distance=1.5cm] {$\boldsymbol{M}_{r}(f)+f$ };
\end{tikzpicture}
  \caption{Residual learning: a building block from \cite{he2015deep} .}
  \label{fig:residual_net}
\end{figure}

In the following, we will further explain the implementation of residual learning in ResNet based on \cite{he2015deep}. The formulation of $\boldsymbol{M}_{r}(f) + f$ can be realized by devising neural networks with “shortcut connections”, which are those connections skipping one or more layers \cite{bishop1995neural,ripley2007pattern,venables2013modern}. In ResNet, the shortcut connections simply perform identity mappings, and their outputs are added to the outputs of the stacked layers as shown in Fig.~\ref{fig:residual_net}. These identity shortcut connections add neither extra parameter nor computational complexity. The entire network can still be trained end-to-end with SGD and can be easily implemented by using common deep learning frameworks (e.g., Caffe \cite{jia2014caffe}, MXNet \cite{chen2015mxnet}, Tensorflow \cite{abadi2016tensorflow} and SPeeDO \cite{zheng2015speedo}) without modifying the solvers.

Formally, a building block of ResNet is defined as:

\begin{equation}
z = \boldsymbol{M}_{r}(f,\{W_{i}\}) + f.
\label{eq:res1}
\end{equation}
Here, $f$ and $z$ are the input and output vectors of the stacked layers, respectively. The function $\boldsymbol{M}_{r}(f,\{W_{i}\})$ is the residual mapping to be learned. For example, there are two layers in Fig.~\ref{fig:residual_net}, and hence the residual mapping for this devised building block is $\boldsymbol{M}_{r} = W_{2}g(W_{1}f)$, where ResNet chooses ReLU as the nonlinear activation function $g$, and the biases are omitted for simplicity. The operation $\boldsymbol{M}_{r} + f$ is performed by using a shortcut connection and then an element-wise addition. The second nonlinearity is employed after the addition (i.e., $g(z)$, see Fig.~\ref{fig:residual_net}). In \cite{he2015deep}, ResNet only employs a residual mapping $\boldsymbol{M}_{r}$ that has two or three layers, while more layers are permitted. 

However, Eq. \ref{eq:res1} assumes that the dimensions of $f$ and $\boldsymbol{M}_{r}$ must be the same. If the dimensions of the input and output channel differ, a linear projection $W_{s}$ should be performed to match the dimensions. For this scenario, a building block turns out to be: 

\begin{equation}
z = \boldsymbol{M}_{r}(f,\{W_{i}\}) + W_{s}f.
\end{equation}

For easy exposition, the above notations concern fully-connected layers, but they are also applicable to convolutional layers. The mapping $\boldsymbol{M}_{r}(f,\{W_{i}\})$ can represent multiple convolutional layers. The element-wise addition is performed on two feature maps, channel by channel.

\section{Transfer Representation Learning}

The success of CNNs relies on not only a good model to capture representations, but also substantial amount of training data to learn representations from.  Unfortunately, in many application domains such as medicine, training data can be scarce, and approaches such as data augmentation are not applicable.  Transfer representation learning is a plausible alternative, which can remedy the insufficient training data issue. 
The common practice of transfer representation learning is to pre-train a CNN on a very large dataset (called source domain) and then to use the pre-trained CNN either as an initialization or a fixed feature extractor for the task of interest  (called target domain) \cite{CS231TL}. 

We use disease diagnosis as the target domain to illustrate the problems of and solutions to the challenges of small data training.
Specifically, we use otitis media (OM) and melanoma\footnote{In our award-winning XPRIZE Tricorder \cite{tricorder_website} device (code name DeepQ), we effectively diagnose twelve conditions, and OM and melanoma are two of them.} as two example diseases.  
The training data available to us are 1) $1,195$ OM images collected by seven otolaryngologists at Cathay General Hospital\footnote{The dataset was used under a strict IRB process.  The dataset was deleted by April 2015 after our experiments had completed.}, Taiwan \cite{shie2014hybrid} and 2) $200$ melanoma images from PH$^2$ dataset \cite{mendoncca2013ph}. 
The source domain from which representations are transfered to our two target diseases is  ImageNet~\cite{deng2009imagenet}, which we have extensively discussed in the prior sections.

What are symptoms or characteristics of OM and melanoma?
OM is any inflammation or infection of the middle ear, and treatment consumes significant medical resources each year \cite{american2004diagnosis}. Several symptoms such as redness, bulging, and tympanic membrane perforation may suggest an OM condition. Color, geometric, and texture descriptors may help in recognizing these symptoms. However, specifying these kinds of features involves a hand-crafted process and therefore requires domain expertise. Often times, human heuristics obtained from domain experts may not be able to capture the most discriminative characteristics, and hence the extracted features cannot achieve high detection accuracy. 
Similarly, melanoma, a deadly skin cancer, is diagnosed based on the widely-used dermoscopic ``ABCD'' rule~\cite{stolzABCD}, where A means asymmetry, B means border, C color, and D different structures. The precise identification of such visual cues relies on experienced dermatologists to articulate. Unfortunately, there are many congruent patterns shared by melanoma and nevus, with skin, hair, and wrinkles often preventing noise-free feature extraction.

Our transfer representation learning experiments consist of the following five steps:

\begin{enumerate}
\item Unsupervised codebook construction: We learned a codebook from ImageNet images, and this codebook construction is ``unsupervised'' with respect to OM and melanoma.
\item Encode OM and melanoma images using the codebook: Each image was encoded into a weighted combination of the pivots in the codebook. The weighting vector is the feature vector of the input image.
\item Supervised learning: Using the transfer-learned feature vectors, we then employed supervised learning to learn two classifiers from the $1,195$ labeled OM instances or $200$ labeled melanoma instances.
\item Feature fusion: We also combined some heuristic features of OM (published in \cite{shie2014hybrid}) and ABCD features of melanoma with features learned via transfer learning.
\item Fine tuning: We further fine-tuned the weights of the CNN using labeled data to improve classification accuracy.
\end{enumerate} 

As we will show in the remainder of this section,  step four does not yield benefit, whereas the other steps are effective in improving diagnosis accuracy.  In other words, these two disease examples demonstrate that features modeled by domain experts or physicians (the model-centric approach) are ineffective. The data-driven approach of big data representation learning combined with small data adaptation is convincingly promising.


\subsection{Method Specifications}
\label{om:method}

We started with unsupervised codebook construction. On the large ImageNet dataset, we learned the representation of these images using AlexNet~\cite{krizhevsky2012imagenet}, presented in Subsection~\ref{cnn:alexnet}. AlexNet contains eight neural network layers. The first five are convolutional and the remaining three are fully-connected. Different hidden layers represent different levels of abstraction concepts. We utilized AlexNet in Caffe~\cite{jia2014caffe} as our foundation to build our encoder to capture generic visual features.

\begin{figure}[h]
  \centering
  \includegraphics[scale=.45]{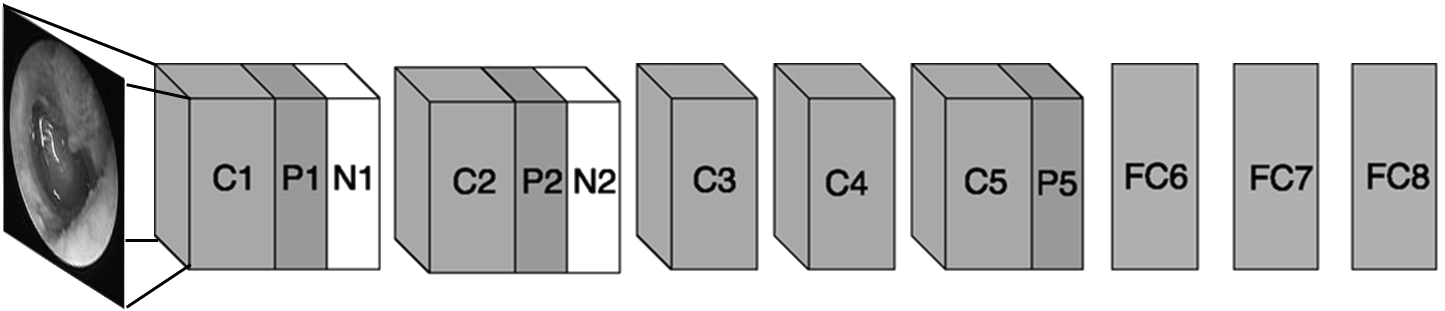}
  \caption{The flowchart of our transfer representation learning algorithm using OM images as example. (Otitis media photo is from ~\cite{OM_website} ) }
  \label{f1}
\end{figure}

For each image input, we obtained a feature vector using the codebook. The information of the image moves from the input layer to the output layer through the inner hidden layers. Each layer is a weighted combination of the previous layer and stands for a feature representation of the input image. Since the computation is hierarchical, higher layers intuitively represent higher concepts. For images, the neurons from lower levels describe rudimentary perceptual elements like edges and corners, whereas the neurons from higher layers represent aspects of objects such as their parts and categories. To capture high-level abstractions, we extracted transfer-learned features of OM and melanoma images from the fifth, sixth and seventh layers, denoted as pool$5$(P$5$), fc$6$ and fc$7$ in Fig.~\ref{f1} respectively. 

\begin{figure}[t]
  \centering
  \includegraphics[scale=.4]{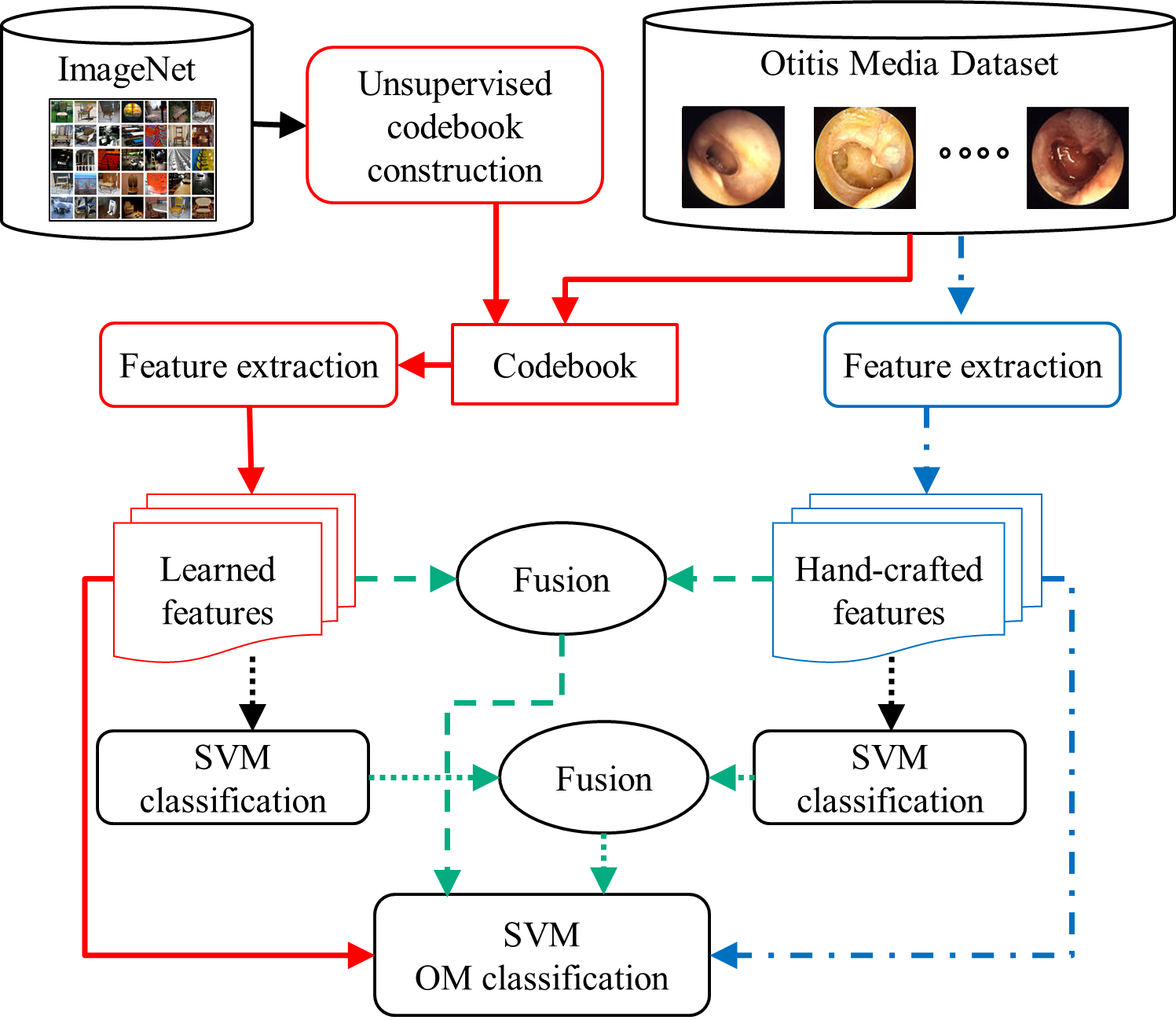}
  \caption{Four classification flows (OM photos are from~\cite{OM_website})}
  \label{f2}
\end{figure}

Once we had transfer-learned feature vectors of the $1,195$ collected OM images and $200$ melanoma images, we performed supervised learning by training a support vector machine (SVM) classifier~\cite{chang2011libsvm}. We chose SVMs to be our model since it is an effective classifier widely used by prior works. Using the same SVM algorithm lets us perform comparisons with the other schemes solely based on feature representation. As usual, we scaled features to the same range and found parameters through cross validation. For fair comparisons with previous OM works, we selected the radial basis function (RBF) kernel.

To further improve classification accuracy, we experimented with two feature fusion schemes, which combine OM features hand-crafted by human heuristics (or model-centric)  in~\cite{shie2014hybrid} and our melanoma heuristic features with features learned from our codebook. In the first scheme, we combined transfer-learned and hand-crafted features to form fusion feature vectors. We then deployed the supervised learning on the fused feature vectors to train an SVM classifier. In the second scheme, we used the two-layer classifier fusion structure proposed in~\cite{shie2014hybrid}. In brief, in the first layer we trained different classifiers based on different feature sets separately. We then combined the outputs from the first layer to train the classifier in the second layer.

Fig.~\ref{f2} summarizes our transfer representation learning approaches using OM images as an example. The top of the figure depicts two feature-learning schemes: the transfer-learned scheme on the left-hand side and the hand-crafted scheme on the right. The solid lines depict how OM or melanoma features are extracted via the transfer-learned codebook, whereas the dashed lines represent the flow of hand-crafted feature extraction. The bottom half of the figure describes two fusion schemes. Whereas the dashed lines illustrate the feature fusion by concatenating two feature sets, the dotted lines show the second fusion scheme at the classifier level. At the bottom of the figure, the four classification flows yield their respective OM-prediction decisions. In order from left to right in the figure are 'transfer-learned features only', 'feature-level fusion', 'classifier-level fusion', and 'hand-crafted features only'.

\subsection{Experimental Results and Discussion}
\label{om:experiment}

Two sets of experiments were conducted to validate our idea. In this subsection, we first report OM classification performance by using our proposed transfer representation learning approach, followed by our melanoma classification performance. Then, we elaborate the correlations between images of ImageNet classes and images of disease classes by using a visualization tool to explain why transfer representation learning works.

For fine-tuning experiments, we performed a $10$-fold cross-validation for OM and a $5$-fold cross-validation for melanoma to train and test our models, so the test data are separated from the training dataset.  We applied data augmentation, including random flip, mirroring, and translation, to all the images.

For the setting of training hyperparameters and network architectures, we used mini-batch gradient descent with a batch size of $64$ examples, learning rate of $0.001$, momentum of $0.9$ and weight decay of $0.0005$. To fine-tune the AlexNet model, we replaced the fc$6$, fc$7$ and fc$8$ layers with three new layers initialized by using a Gaussian distribution with a mean of $0$ and a std of $0.01$. During the training process, the learning rates of those new layers were ten times greater than that of the other layers.

\subsubsection{Results of transfer representation learning for OM}

Our $1,195$ OM image dataset encompasses almost all OM diagnostic categories: normal; AOM: hyperemic stage, suppurative stage, ear drum perforation, subacute/resolution stage, bullous myringitis, barotrauma; OME: with effusion, resolution stage (retracted); COM: simple perforation, active infection.
Table~\ref{OM_table} compares OM classification results for different feature representations. All experiments were conducted using $10$-fold SVM classification. The measures of results reflect the discrimination capability of the features.

The first two rows in Table~\ref{OM_table} show the results of human-heuristic methods (hand-crafted), followed by our proposed transfer-learned approach. The eardrum segmentation, denoted as ‘seg’, identifies the eardrum by removing OM-irrelevant information such as ear canal and earwax from the OM images~\cite{shie2014hybrid}. The best accuracy achieved by using human-heuristic methods is around $80\%$. With segmentation (the first row), the accuracy improves $3\%$ over that without segmentation (the second row).

Rows three to eight show results of applying transfer representation learning. All results outperform the results shown in rows one and two, suggesting that the features learned from transfer learning are superior to that of human-crafted ones.

Interestingly, segmentation does not help improve accuracy for learning representation via transfer learning. This indicates that the transfer-learned feature set is not only more discriminative but also more robust. Among three transfer-learning layer choices (layer five (pool$5$), layer six (fc$6$) and layer seven (fc$7$)), fc$6$ yields slightly better prediction accuracy for OM. We believe that fc$6$ provides features that are more general or fundamental to transfer to a novel domain than pool$5$ and fc$7$ do. (Section~\ref{sec-qual} presents qualitative evaluation and explains why for OM fc$6$ is ideal.)

We also directly used the $1,195$ OM images to train a new AlexNet model.The resulting classification accuracy was only $71.8\%$, much lower than applying transfer representation learning. This
result confirms our hypothesis that even though CNN is a good model, with merely $1,195$ OM images (without the ImageNet images to facilitate feature learning), it cannot learn discriminative features. 

Two fusion methods, combining both hand-crafted and transfer learning features, achieved a slightly higher OM-prediction F1-score ($0.9$ over $0.895$) than using transfer-learned features only. This statistically insignificant improvement suggests that hand-crafted features do not provide much help.

Finally, we used OM data to fine-tune the AlexNet model, which achieves the highest accuracy. For fine-tuning, we replaced the original fc$6$, fc$7$ and fc$8$ layers with the new ones and used OM data to train the whole network without freezing any parameters. In this way, the leaned features can be refined and are thus more aligned to the targeted task.  This result attests that the ability to adapt representations to data is a critical characteristic that makes deep learning superior to the other learning algorithms.

\begin{table}[t]
  \caption{OM classification experimental results}
  \newcounter{magicrownumbers}
 \newcommand\rownumber{\stepcounter{magicrownumbers}\arabic{magicrownumbers}}
  \label{OM_table}
  \renewcommand\arraystretch{1.2}
  \centering
  \begin{tabular}{l|l|llll}
    \toprule
    & Method     & Accuracy(std)   & Sensitivity  & Specificity  &  F\_1-Score\\
    \midrule
    \rownumber & Heuristic w/ seg        & $80.11\%(18.8)$ & $83.33\%$  & $75.66\%$  &  $0.822$ \\ 
    \rownumber & Heuristic w/o seg       & $76.19\%(17.8)$ & $79.38\%$  & $71.74\%$  &  $0.79$  \\
    \hline
    \rownumber & Transfer w/ seg (pool$5$) & $87.86\%(3.62)$ & $89.72\%$  & $86.26\%$  &  $0.89$  \\
    \rownumber & Transfer w/o seg (pool$5$)& $88.37\%(3.41)$ & $89.16\%$  & $87.08\%$  &  $0.894$ \\
    \rownumber & Transfer w/ seg (fc$6$)   & $87.58\%(3.45)$ & $89.33\%$  & $85.04\%$  &  $0.887$  \\
    \rownumber & Transfer w/o seg (fc$6$)   & {\bf $88.50\%(3.45)$} & $89.63\%$  & $86.90\%$  &  $0.895$ \\
    \rownumber & Transfer w/ seg (fc$7$)    & $85.60\%(3.45)$ & $87.50\%$  & $82.70\%$  &  $0.869$ \\
    \rownumber & Transfer w/o seg (fc$7$)   & $86.90\%(3.45)$ & $88.50\%$  & $84.90\%$  &  $0.879$ \\
    \hline
    \rownumber & Feature fusion           & $89.22\%(1.94)$ & $90.08\%$  & $87.81\%$  &  $0.90$  \\   
    \rownumber & Classifier fusion        & {\bf $89.87\%(4.43)$} & $89.54\%$  & $90.20\%$  &  $0.898$  \\
    \rownumber & Fine-tune                & {\bf $90.96\%(0.65)$} & $91.32\%$  & $90.20\%$  &  $0.917$ \\    
    \hline
    \bottomrule
  \end{tabular}
\end{table}

\subsubsection{Results of transfer representation learning for Melanoma}

We performed experiments on the PH$^2$ dataset whose dermoscopic images were obtained at the Dermatology Service of Hospital Pedro Hispano (Matosinhos, Portugal) under the same conditions through the Tuebinger Mole Analyzer system using a magnification of $20$x. The assessment of each label was performed by an expert dermatologist.

Table~\ref{Melnoma_table} compares melanoma classification results for different feature representations. In Table~\ref{Melnoma_table}, all the experiments except for the last two were conducted by using $5$-fold SVM classification. The last experiment involved fine-tuning, which was implemented and evaluated by using Caffe. We also performed data augmentation to balance the PH$^2$ dataset ($160$ normal images and $40$ melanoma images) . 

Unlike OM, we found the low-level features to be more effective in classifying melanoma. Among three transfer-learning layer choices, pool$5$ yields a more robust prediction accuracy than the other layers do for melanoma. The deeper the layer is, the worse the accuracy becomes.  We believe that pool$5$ provides low-level features that are suitable for delineating texture patterns that depict characteristics of melanoma. 

Rows three and seven show that the accuracy of transferred features is as good as that of the ABCD rule method with expert segmentation. These results reflect that deep transferred features are robust to noise such as hair or artifacts.

We used melanoma data to fine-tune the AlexNet model and obtained the best accuracy $92.81\%$ since all network parameters are refined to fit the target task by employing back propagation. We also compared our result with the cutting-edge method, which reported $98\%$ sensitivity  and $90\%$ specificity on PH$^2$ ~\cite{barata2015melanoma}. Their method requires preprocessing such as manual lesion segmentation to obtain ``clean'' data. In contrast, we utilized raw images without conducting any heuristic-based preprocessing. Thus, deep transfer learning can identify features in an unsupervised way to achieve as good classification accuracy as those features identified by domain experts.

\begin{table}[t]
  \caption{Melanoma classification experimental results}
  \newcounter{magicrownumberss}
 \newcommand\rownumber{\stepcounter{magicrownumberss}\arabic{magicrownumberss}}
  \label{Melnoma_table}
  \renewcommand\arraystretch{1.2}
  \centering
  \begin{tabular}{l|l|lllll}
    \toprule
    & Method     & Accuracy(std)   & Sensitivity  & Specificity  &  F\_1-Score\\
    \midrule
    \rownumber & ABCD rule w/ auto seg        & $84.38\%(13.02)$ & $85.63\%$  & $83.13\%$  &  $0.8512$ \\ 
    \rownumber & ABCD rule w/ manual seg      & $89.06\%(9.87)$ & $90.63\%$  & $87.50\%$  &  $0.9052$  \\
    \rownumber & Transfer w/o seg (pool$5$)     & \textbf{$89.06\%(10.23)$} & $92.50\%$  & $85.63\%$  &  $0.9082$ \\
    \rownumber & Transfer w/o seg (fc$6$)       & $85.31\%(11.43)$ & $83.13\%$  & $87.50\%$  &  $0.8686$ \\
    \rownumber & Transfer w/o seg (fc$7$)       & $79.83\%(14.27)$ & $84.38\%$  & $74.38\%$  &  $0.8379$ \\
    \rownumber & Feature fusion               & $90.0\%(9.68)$ & $92.5\%$  & $87.5\%$  &  $0.9157$  \\   
    \rownumber & Fine-tune                    & {\bf $92.81\%(4.69)$} & $95.0\%$  & $90.63\%$  &  $0.93$  \\
 \hline   
 \bottomrule
  \end{tabular}
\end{table}

\subsubsection{Qualitative Evaluation - Visualization}
\label{sec-qual}

In order to investigate what kinds of features are transferred or borrowed from the ImageNet dataset, we utilized a visualization tool to perform qualitative evaluation. Specifically, we used an attribute selection method, SVMAttributeEval~\cite{guyon2002gene} with Ranker search, to identify the most important features for recognizing OM and melanoma. Second, we mapped these important features back to their respective codebook and used the visualization tool from ~\citet{yosinski2015understanding} to find the top ImageNet classes causing the high value of these features. By observing the common visual appearances shared by the images of the disease classes and the retrieved top ImageNet classes, we were able to infer the transferred features.

\begin{figure}[h]
  \centering
  \includegraphics[scale=0.5]{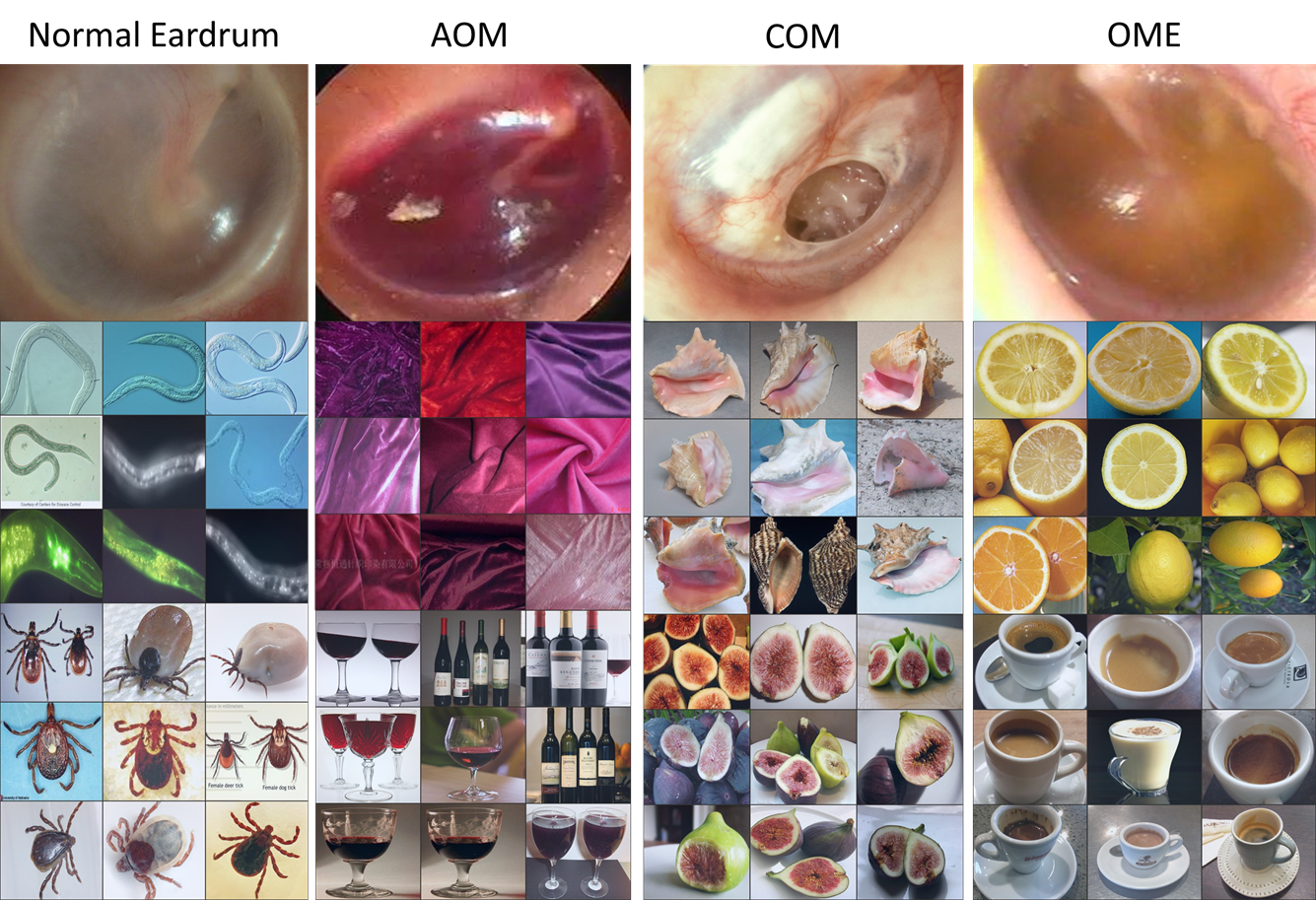}
  \caption{The visualization of helpful features from different classes corresponding to different OM symptoms(from left to right: Normal eardrum, AOM, COM, OME)}
  \label{f3}
\end{figure}

Fig.~\ref{f3} 
demonstrates the qualitative analyses of four different cases: the Normal eardrum, acute Otitis Media (AOM), Chronic Otitis Media (COM) and Otitis Media with Effusion (OME), which we will now proceed to explain in turn.  First, the normal eardrum, nematode and ticks are all similarly almost gray with a certain degree of transparency, features that are hard to capture with only hand-crafted methods. Second, AOM, purple-red cloth and red wine have red colors as an obvious common attribute. Third, COM and seashells are both commonly identified by a calcified eardrum. Fourth, OME, oranges, and coffee all seem to share similar colors. Here, transfer learning works to detect OM in an analogous fashion to how {\em explicit similes} are used in language to clarify meaning. The purpose of a simile is to provide information about one object by comparing it to something with which one is more familiar. For instance, if a doctor says that OM displays redness and certain textures, a patient may not be able to comprehend the doctor’s description exactly. However, if the doctor explains that OM presents with an appearance similar to that of a seashell, red wine, orange, or coffee colors, the patient is conceivably able to envision the appearance of OM at a much more precise level. At level fc$6$, transfer representation learning works like finding similes that can help explain OM using the representations learned in the source domain (ImageNet). 

\begin{figure}[h]
  \centering
  \includegraphics[scale=0.35]{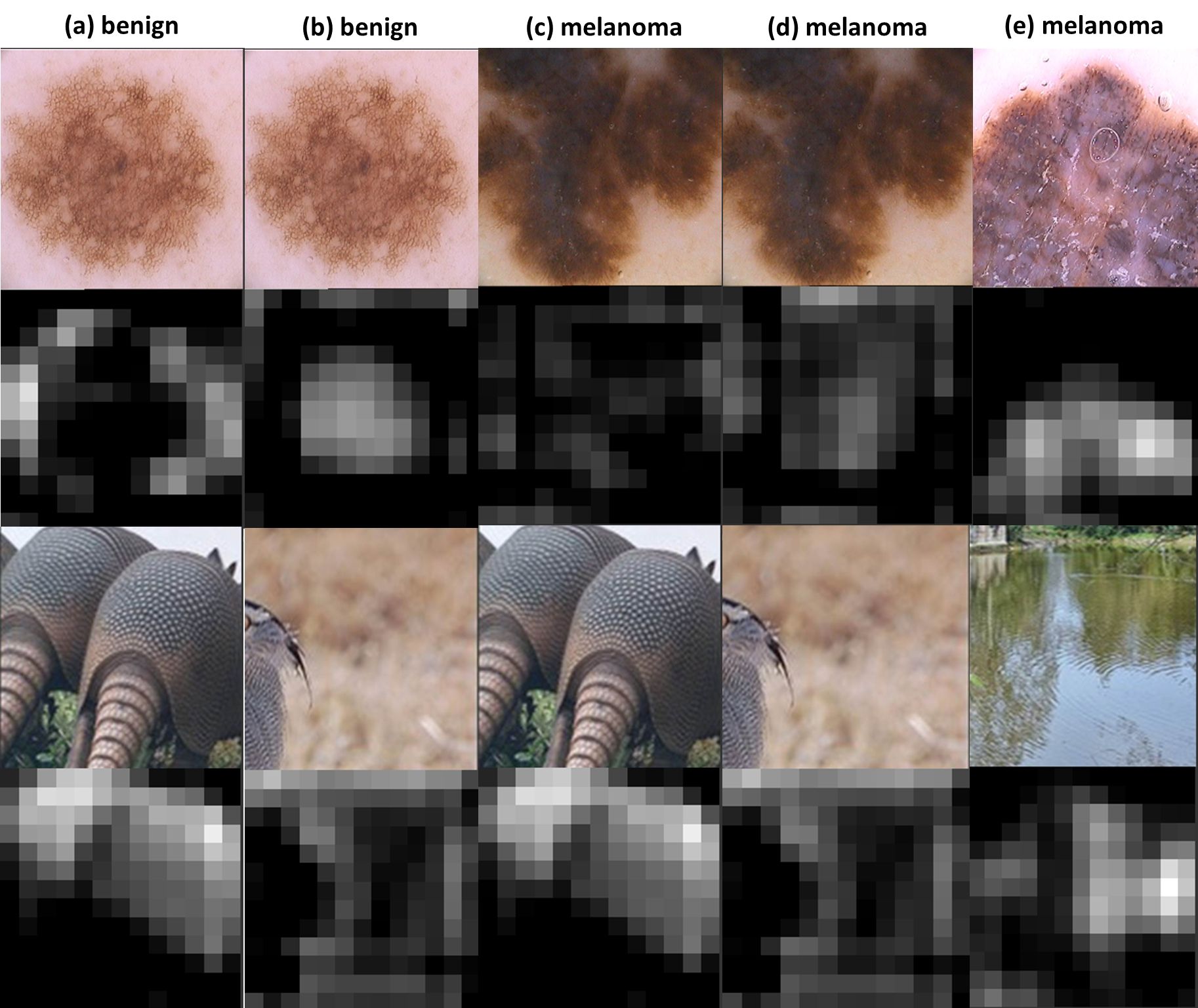}
  \caption{The visualization of helpful features for different patterns.}
  \label{f4}
\end{figure}

To illustrate melanoma detection, 
Fig.~\ref{f4} presents visualization for three images, one benign nevus and two melanoma images.  Columns (a) and (b) in Fig.~\ref{f4} show the same benign nevus image and its two most representative features.  Columns (c) and (d) show the same melanoma image and its representative features.  Column (e) displays a melanoma image with different visual characteristics. 

The top row of the figure shows three images inputted into the visualization tool. The second row reflects the activation maps, which we derived by employing C$5$ kernels on the input images. The larger the activation value used, the brighter the corresponding pixel returned. Using the respective kernels, the third row demonstrates the maximal activation map each image can reach ~\cite{yosinski2015understanding}. We can thus conclude that each kernel is learned to detect specific features in order to capture the appearance of the images. 

The classification of melanoma contrasts sharply with the classification of OM. We can exploit distinct visual features to classify different OM classes. However, melanoma and benign nevi share very similar textures, as melanoma evolves from benign nevi. Moreover, melanoma often has atypical textures and presents in various colors. 

Among the three kinds of features in Fig.~\ref{f4}, two of them are shared between benign nevi and melanoma. One is learned from the Cingulata shown in columns (a) and (c). According to columns
(a)	and (c), we can see that nevi, melanoma and Cingulata share a pigment network and lattice pattern; the nevi in column (a) has a surrounding lattice pattern that causes a bright outline around the activation map in the second row. Other shared features are learned from the dirt images in columns (b) and (d); we can observe that nevi, melanoma and dirt have similar colors and textures. For example, the nevi in column (b) is dirt-like in its interior, thus leading to the light circle in the second row. Column (e) shows one of the critical melanoma characteristics --- a “Blue-Whitish Veil” --- which shares the same texture as that of the sparkle of sunlight on the lake (the third row of column (e)).

In the case of detecting melanoma versus benign nevus, effective representations of the diseases from higher-level visual characteristics cannot be found from the source domain. Instead, the most effective representations are only transferrable at a lower-level of the CNN. We believe that if the source domain can add substantial images of dirt and texture-rich objects, the effect of explicit similes may be utilized at a higher level of the CNN.

\subsection{Observations on Transfer Learning}
\label{om:conclusion}

In summary, this section demonstrates that transfer representation learning can potentially remedy two major challenges of medical image analysis: labeled data scarcity and medical domain knowledge shortage. Representations of OM and melanoma images can be effectively learned via transfer learning.

The transfer learned features achieve accuracy $90.96\%$ ($91.32\%$ in sensitivity and $90.20\%$ in specificity) for OM and $92.81\%$ ($95.0\%$ in sensitivity and $90.63\%$ in specificity) for melanoma, achieving an improvement in disease-detection accuracy over the feature extraction instructed by domain experts. Moreover, our algorithms do not require manual data cleaning beforehand, and the preliminary diagnosis of OM and melanoma can be derived without aid from doctors. Therefore, automatic disease diagnosis systems, which hold the potential to help people lacking in access to medical resources, are developmentally possible.

\section{Concluding Remarks}

Deep learning owes its success to three key factors: scale of data, enhanced models to learn representations from data, and scale of computation.  This chapter presented the importance of the data-driven approach to learn good representations from both big data and small data.  

In terms of big data, it has been widely accepted in the research community that the more data the better for both representation and classification improvement.  The question is then how to learn representations from big data, and how to perform representation learning when data is scarce.  We addressed the first question by presenting CNN model enhancements in the aspects of representation, optimization, and generalization.  To address the small data challenge, we showed 
transfer representation learning to be effective. 
Transfer representation learning transfers the learned representation from a source domain where abundant training data is available to a target domain where training data is scarce.  Transfer representation learning gave the OM and melanoma diagnosis modules of our XPRIZE Tricorder device (which finished $2^{nd}$ out of $310$ competing teams~\cite{tricorder_website}) a significant boost in diagnosis accuracy.   

Our experiments on transfer learning provided three important insights on representation learning.  

\begin{enumerate}
\item Low-level representations can be shared. Low-level perceptual features such as edges, corners, colors, and textures can be borrowed from some source domains where training data are abundant.  After all, low-level representations are similar despite different high-level semantics.

\item Middle-level representations can be correlated.  Analogous to explicit similes used in language, an object in the target domain can be ``represented'' or ``explained'' by some source domain features. In our OM visualization, we observed that a positive OM may display appearances similar to that of a seashell, red wine, oranges, or coffee colors --- features learned and transferred from the ImageNet source domain. 

\item Representations can adapt to a target domain. Even though, in the small data training situations, the amount of data is insufficient to learn effective representations by itself, given representations learned from some big-data source domains, the small data of the target domain can be used to align (e.g., re-weight) the representations learned from the source domains to adapt to the target domain.
\end{enumerate}


Finally, we suggest further readings, which provide further details on improving scale of data and taking advantage of scale of computation to speed up representation learning as follows.

For scale of data, please refer to \cite{chang2011foundationsch2}, which was originally submitted to Transactions of IEEE as an invited paper in 2010 but was rejected because a CNN pioneer did not consider scale of data to be a critical factor. (Two US patents on feature extraction~\cite{Patent9547914} and objection recognition~\cite{Patent8798375} using a hybrid approach of deep learning (model-centric) and big data (data-driven) were submitted in 2011 and granted in 2017 and 2014, respectively.) 
Two years later, AlexNet spoke volumes in support of the importance scale of data. For recent representative works in increasing data scale via synthetic data or unlabeled data, please consult \cite{jaderberg2014synthetic,peng2015learning,rogez2016mocap,shrivastava2016learning,sixt2016rendergan}.

For an introduction to the computation of neural network models, please refer to \cite{dally2015tutorial}. For available deep learning frameworks, please refer to Torch \cite{collobert2011torch7}, Caffe \cite{jia2014caffe}, MXNet \cite{chen2015mxnet}, Theano \cite{al2016theano}, and TensorFlow \cite{abadi2016tensorflow}. For scale of computation, please consult pioneering work at Google \cite{GoogleBigData2008}, and subsequent efforts on accelerating deep learning such as DistBelief \cite{dean2012large}, Adam \cite{chilimbi2014project}, and SpeeDO \cite{zheng2015speedo}.

\bibliographystyle{plainnat}
\bibliography{ref}



\end{document}